\pdfoutput=1

\documentclass[11pt]{article}

\usepackage[final]{acl}

\usepackage{graphics}
\usepackage{enumitem}
\usepackage{makecell}
\usepackage{graphicx}
\usepackage{fvextra}
\usepackage{spverbatim}

\usepackage{times}
\usepackage{latexsym}
\usepackage{multirow}

\usepackage{graphicx}
\usepackage{tikz}

\usepackage{listings}

\usepackage[T1]{fontenc}

\usepackage[utf8]{inputenc}

\usepackage{microtype}
\usepackage{booktabs}
\usepackage{tablefootnote}

\usepackage{inconsolata}

\usepackage{amsmath}

\title{SmurfCat at SemEval-2024 Task 6: Leveraging Synthetic Data for Hallucination Detection}

\author{
\textbf{Elisei Rykov\textsuperscript{1,2}}, \hspace{2pt} \textbf{Yana Shishkina\textsuperscript{2,3}}, \textbf{Kseniia Petrushina\textsuperscript{1,4}}\hspace{2pt},\\ \textbf{Kseniia Titova\textsuperscript{1,5}}, \textbf{Sergey Petrakov\textsuperscript{1}}, \textbf{and}
\textbf{Alexander Panchenko\textsuperscript{1,6}} \\
\textsuperscript{1}Skolkovo Institute of Science and Technology,
\textsuperscript{2}Tinkoff,\\
\textsuperscript{3}HSE University,
\textsuperscript{4}Moscow Institute of Physics and Technology,
\textsuperscript{5}MTS AI,
\textsuperscript{6}AIRI\\
\href{mailto:e.rykov@tinkoff.ai}{\small \textsf{\{e.rykov, y.a.shishkina\}@tinkoff.ai}},
\href{mailto:kseniia.petrushina@skol.tech}{\small \textsf{\{kseniia.petrushina, kseniia.titova, sergey.petrakov, a.panchenko\}@skol.tech}}
}

\begin{document}
\maketitle
\begin{abstract}
In this paper, we present our novel systems developed for the SemEval-2024 hallucination detection task. Our investigation spans a range of strategies to compare model predictions with reference standards, encompassing diverse baselines, the refinement of pre-trained encoders through supervised learning, and an ensemble approaches utilizing several high-performing models. Through these explorations, we introduce three distinct methods that exhibit strong performance metrics. To amplify our training data, we generate additional training samples from unlabelled training subset. Furthermore, we provide a detailed comparative analysis of our approaches. Notably, our premier method achieved a commendable 9th place in the competition's model-agnostic track and 17th place in model-aware track, highlighting its effectiveness and potential.
\end{abstract}

\section{Introduction}

Large language models are proficient in generating human-like text across various styles. However, even the most advanced models can produce hallucinations, leading users to question their reliability. There are two primary types of hallucinations: factuality hallucinations, which involve the generation of content that deviates from actual facts, and faithfulness hallucinations, when the model fails to solve tasks correctly following specific instructions \cite{huang2023survey}. 

The SemEval 2024 Shared-task on Hallucinations and Related Observable Overgeneration Mistakes \cite{mickus-etal-2024-semeval} has integrated both types into three tasks. The Definition Modeling task (DM) focused on fact-related hallucinations by challenging models to generate contextually relevant word definitions. Both the Machine Translation (MT) and Paraphrase Generation (PG) tasks included faithfulness hallucinations, with models asked to produce translations or paraphrases for given sentences. Evaluation labelled datasets for these tasks were provided and the training dataset consisted only of source sentences and model generations, without corresponding labels.

Motivated by the lack of annotated resources and the efficacy of other language models trained on synthetic data, we developed two synthetic datasets that replicate the targeted domain. First, we collected data through a proprietary GPT-4  model \cite{OpenAI_GPT4_2023}, but our methods trained on the achieved data did not yield the desired results as prompt engineering made maintaining the domain challenging. As a second approach, we trained LLaMA2-7b \cite{touvron2023llama} adapters using a small set of annotated examples and applied them to the unlabeled training data. This method proved to be a more effective form of in-domain data augmentation.

While the competition was run on two tracks, we focus mainly on the model-agnostic track. In our methods we utilized the most effective models with varied sizes and architectures, which we had evaluated beforehand. Our experiments involved fine-tuning a pre-trained embedding model, repurposing it to function as a binary classifier across a number of open-source datasets, including our synthetic sets. We also experimented with a promising method for evaluating paraphrases by modifying its design and fine-tuning the model on different data. Finally, we tested different combinations of the highest-performing approaches in an ensemble setting. Generated synthetic data and code published on GitHub\footnote{\url{https://github.com/s-nlp/shroom}}.

\section{Related work} \label{sec:related-work}
In the field of text representation, the E5 \cite{wang2022text} family represents a group of cutting-edge sentence embedding models trained through contrastive methods. The E5-Mistral\footnote{\url{https://huggingface.co/intfloat/e5-mistral-7b-instruct}} model, a powerful embedding model that has been fine-tuned on a selection of annotated data, is currently recognized as the leading open-source model by the Multitask Text Embedding Benchmark \cite{muennighoff2022mteb}. 

Vectara's \textit{hallucination\_detection\_model}\footnote{\url{https://huggingface.co/vectara/hallucination_evaluation_model}} is a fine-tuned DeBERTa focused on summarization datasets that includes annotations for factual consistency. TrueTeacher \cite{gekhman-etal-2023-trueteacher} is a family of models and an associated dataset designed for evaluating factual consistency. The dataset was created by first fine-tuning various-sized T5 models on summarization tasks. These models were then employed to generate hypotheses, which were subsequently automatically annotated using a 540B Large Language Model (LLM). This annotated dataset was then utilized to train multiple models to assess factual consistency.

The Mutual Implication Score (MIS) \cite{babakov-etal-2022-large} is a metric devised for evaluating the quality of text style transfer and paraphrasing systems, grounding its assessment on content similarity between the prediction and the reference text. It leverages a RoBERTa-NLI \cite{nie-etal-2020-adversarial} model that has been fine-tuned and incorporates it into an architecture that processes two input texts sequentially in both forward and reverse directions. The final hidden states from these two passes are merged and forwarded to a classification head to determine the MIS score. Initially, the MIS metric was trained using the Quora Question Pairs dataset (QQP) \cite{Sharma2019NaturalLU}.

SimCSE (Similarity-based Contrastive Self-supervised Learning) \cite{gao2021simcse} is a self-supervised learning method for text embeddings. It is used for creating embeddings of text data that are semantically meaningful and can be used in various downstream tasks. It involves training a neural network to maximize the similarity between embeddings of similar sentences and minimize the similarity  between embeddings of dissimilar sentences. LaBSE (Language-agnostic BERT Sentence Embedding) \cite{feng2020languageagnostic} is a method for generating multilingual sentence embeddings using the BERT architecture.

Other metrics for evaluating content preservation, such as BLEU (Bilingual Evaluation Understudy) \cite{papineni2002bleu}, CHRF (Character n-gram F-score) \cite{popovic2015chrf}, METEOR (Metric for Evaluation of Translation with Explicit ORdering) \cite{banerjee2005meteor}, and BLEURT (Bilingual Evaluation Understudy for Ranking and Tuning) \cite{sellam2020bleurt}, also stand out. BLEU utilizes a modified unigram precision score, CHRF evaluates the quality of machine translation by comparing character n-grams in candidate translations against reference translations to compute an F-score, METEOR calculates the harmonic mean of precision and recall at the single-word level, and BLEURT employs a fine-tuned BERT model in a cross-encoder setup, using synthetic data to assess semantic similarity.

\begin{figure}
\centering
\includegraphics[width=\columnwidth]{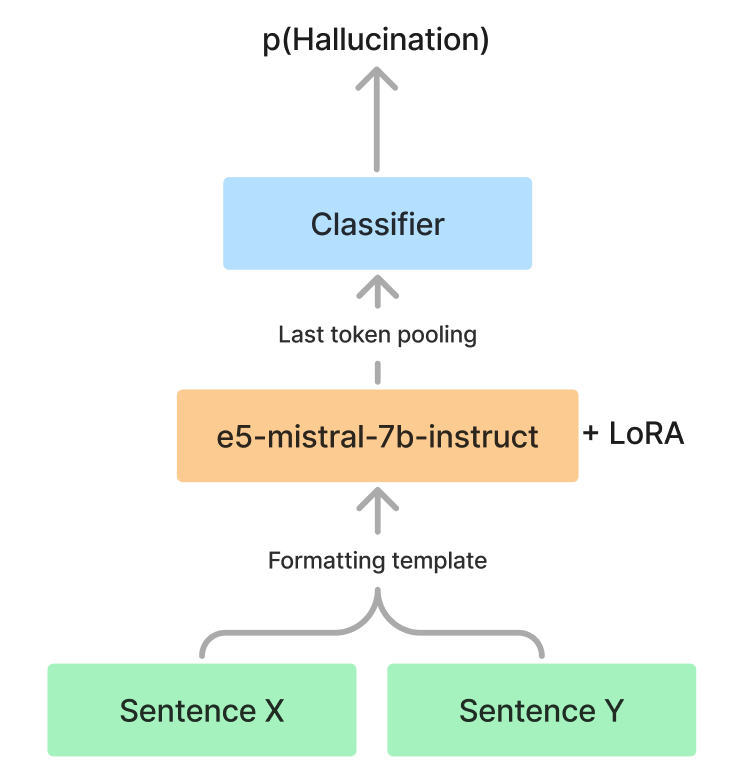}
\caption{Classifier architecture when using synthetic data.}
\label{figure:e5-mistral-classifier-architecture}
\end{figure}

\section{Data}
\subsection{Existing datasets}
The QQP dataset consists of pairs of questions from the Quora forum. For each pair, it is indicated whether the questions are paraphrases, i.e. they ask about the same thing. PAWS \cite{zhang-etal-2019-paws} is a paraphrase detection dataset that contains complex cases with both paraphrase and non-paraphrase samples that have high lexical overlap. 

We postulated that other pre-existing datasets, such as QQP and PAWS, might exhibit particular biases due to their distinct task domains (for instance, QQP dataset includes only questions). To mitigate this potential issue, we generated synthetic data taking unlabeled training samples as starting points.

Our experimentation with synthetic data creation was divided into two main approaches: the first involved training LoRA \cite{hu2022lora} adapters for the LLaMA2-7b model using the annotated data derived from the validation set. The second approach involved the generation of both correct and incorrect hypotheses by employing GPT-4 and specific prompts.

All tasks were distilled down to the paraphrase evaluation task. Consequently, we only used targets (sources for paraphrase generation) and hypotheses as inputs for the models.

\subsection{LLaMA2-7b adapter}
We trained 6 LoRA adapters, pairing them to specialize in either generating hallucinations or producing correct responses for each task. Due to the limited amount of labeled data, we made use of model's ability of in-context learning by prepending samples with instructions: \textit{Paraphrase} for non-hallucinations and \textit{Provide an incorrect paraphrase} for hallucinations. The number of samples for each adapter is shown in Table \ref{tab:llama-data-amount}.

\begin{table}
\small
\begin{center}
\begin{tabular}{@{}llll@{}}
\toprule
                  & \textbf{DM} & \textbf{MT} & \textbf{PG} \\ \midrule
Not Hallucination & 188         & 211         & 132         \\
Hallucination     & 175         & 179         & 132         \\ \midrule
\textbf{Total}    & 363         & 390         & 264         \\ \bottomrule
\end{tabular}
\caption{Adapter train sample sizes.}
\label{tab:llama-data-amount}
\end{center}
\end{table}

Training and generation hyperparameters are displayed in Table \ref{tab:llama-train-params}. For each task and label we manually selected the best epoch by analyzing a small set of generated samples. These checkpoints were further employed to synthesize hypotheses for their task's training set. A small sample of the generated data using LLaMA2-7b adapter is provided in the Appendix \ref{sec:appendix-synthetic-data}.

\begin{table}
\small
\begin{center}
\begin{tabular}{@{}lll@{}}
\toprule
\textbf{Stage}     & \textbf{Hyperparameter} & \textbf{Value} \\ \midrule
\textbf{Training}  & lr                      & 4e-4           \\
                   & warmup\_steps           & 1              \\
                   & optimizer               & AdamW          \\
                   & scheduler               & linear         \\
                   & LoRA alpha              & 16             \\
                   & LoRA dropout            & 0.05           \\
                   & LoRA r                  & 16             \\
                   & batch size              & 32             \\ \midrule
\textbf{Inference} & num\_beams              & 3              \\
                   & do\_sample              & true           \\
                   & repetition\_penalty     & 1.2            \\
                   & top\_k                  & 50             \\
                   & max\_new\_tokens        & 512            \\ \bottomrule
\end{tabular}
\caption{Training and inference hyperparameters for LoRA adapters.}
\label{tab:llama-train-params}
\end{center}
\end{table}

\subsection{GPT-4 prompting} In addition, we created two distinct prompts for the PG task. In these prompts, we directed GPT-4 to generate a paraphrase of a source sentence extracted from an unlabeled training sample. The nature of the paraphrase, whether it should contain hallucinations and overgeneration errors or not, was determined by the specific prompt we used.

We enriched the prompt structure for few-shot learning purposes, incorporating several illustrative examples drawn from both the validation and trial data splits. Alongside each \textit{incorrect} example, we included an explanation to clarify why the provided hypothesis did not meet the criteria.

Moreover, we tasked GPT-4 to execute its reasoning step-by-step: to iterate through several examples with accompanying explanations, and, by leveraging those explanations, to discern and select the most suitable paraphrase.

We utilized the \textit{gpt-4-1106-preview} model, adhering to the default generation parameters stipulated by the OpenAI API service.


\subsection{Data filtration}
In the process of evaluating the synthetic data we generated, we encountered multiple issues that necessitated an extra layer of filtering:
\setlist{nolistsep}
\begin{itemize}[noitemsep]
    \item A number of the samples produced by the LLaMA2-7B model were excessively lengthy, containing up to 1024 tokens.
    \item The labeling of samples by the LLaMA2-7B as \textit{Hallucination} was frequently incorrect. Samples designated as hallucinations were often devoid of any such content, and conversely, non-hallucination samples sometimes contained hallucinations.
    \item A peculiar pattern was observed in the DM task generations from LLaMA2-7B, where more than 9,000 samples started with the word \textit{any} or \textit{anything} denoting a biased starting point which may impact the diversity and neutrality required for effective training.
    \item In the subsets of synthetic data generated by GPT-4 and labeled as \textit{Not Hallucination} the resulting examples were deemed too straightforward, potentially leading to a training dataset that cannot robustly challenge and thereby improve the model's discriminatory capabilities.
\end{itemize}

To tackle the identified issues with the synthetic data, we adopted a systematic filtering methodology. We began by eliminating any hypothesis that exceeded a length of 200 tokens, ensuring the data remained succinct. For the samples that started with \textit{any} or \textit{anything}, we decided to limit the number to 500 to minimize bias.

\begin{table*}
\small
\begin{center}
\begin{tabular}{@{}lllll@{}}
\toprule
\textbf{Source method}     & \textbf{Task}       & \textbf{Label}    & \textbf{\# before filtering} & \textbf{\# after filtering} \\ \midrule
\multirow{6}{*}{LLaMA2-7B} & \multirow{2}{*}{MT} & Hallucination     & 18 093                       &        7 758                     \\
                           &                     & Not Hallucination & 17 056                       &        3 572                   \\ \cmidrule(l){2-5} 
                           & \multirow{2}{*}{PG} & Hallucination     & 13 961                       &        2 839                     \\
                           &                     & Not Hallucination & 14 928                       &        3 952                     \\ \cmidrule(l){2-5} 
                           & \multirow{2}{*}{DM} & Hallucination     & 19 224                       &        5 939                    \\
                           &                     & Not Hallucination & 20 000                       &        12 032                 \\ \midrule
\multirow{2}{*}{GPT-4}     & \multirow{2}{*}{PG} & Hallucination     & 7 439                        &         -                 \\
                           &                     & Not Hallucination & 6 279                        &         -                \\ \bottomrule
\end{tabular}
\caption{The number of samples in the synthetic datasets. No filtering was performed for GPT-4.}
\label{tab:synt-data-stats}
\end{center}
\end{table*}

With the aim of refining the data quality, we then annotated all the synthetic samples using MIS. We set specific thresholds for these MIS scores to filter the data further. In the subset containing hallucinations, we removed samples that had a score lower than 0.1 or higher than 0.5. For non-hallucinated samples, we only retained those with a score between 0.7 and 0.9. These score ranges were established empirically to ensure a balance between discernibility and ambiguity in both the hallucinated and non-hallucinated examples.

The number of samples generated using both synthetic methods, before and after the filtering stage, is given in Table \ref{tab:synt-data-stats}. After generating the synthetic data, we performed several experiments with different combinations of synthetic data.

\section{Methods}

\subsection{Black-box baselines}
First, we started with an assessment of various baseline models that are detailed in Section \ref{sec:related-work}, including a new addition, GPT-4. These baseline models were utilized as-is, in a \textit{black-box} fashion, without any further fine-tuning specifically for our tasks.

For all models other than GPT-4, we employed the inference code available on the official HuggingFace Hub pages. For GPT-4, we created specific prompts for each task. Within these prompts, we instructed GPT-4 to methodically process the information and ascertain the presence of hallucinations within the sample. We provided all pertinent data (source, hypothesis, and, when available, target) within the prompt. It is important to note that the collection and evaluation of predictions were conducted strictly within the model-aware track. The prompt is available in Appendix \ref{sec:appendix-gpt4-prompt-annot}.
 
\subsection{SFT E5-Mistral}
The obtained synthetic data was used to fine-tune the E5-Mistral model on our domain. In our experiments, we adjusted the data inputs by adding or omitting certain subsets of synthetic data to create the final blend used for training. The choice of the E5-Mistral model as the foundation for our work was based on its superior performance compared to other models.

The design of our classifier is depicted in Figure \ref{figure:e5-mistral-classifier-architecture}. In simple terms, we prepare two sample sentences with a specific format and input them into an model with LoRA. Afterwards, we obtain the embedding of the last token and pass it to the classification head.

\subsection{Mutual Implication Score}
\label{subsection:mis}
In this setup we experimented with some improvements to the original Mutual Implication Score model architecture. Even though MIS was already trained on a large amount of paraphrase detection data, QQP dataset biased to the questions. Therefore, we thought that we can fine-tune it to decrease this bias.

\begin{table}
\small
\begin{center}
\begin{tabular}{@{}ll@{}}
\toprule
\textbf{Hyperparameter} & \textbf{Value} \\ \midrule
lr                      & 1e-4           \\
lr scheduler            & constant       \\
optimizer               & AdamW          \\
batch size              & 32             \\ \bottomrule
\end{tabular}
\caption{Training hyperparameters for MIS experiments.}
  \label{tab:def-args}
\end{center}
\end{table}

In Table~\ref{tab:def-args} we present default training hyperparameters used for  experiments with MIS. Unless stated otherwise, we chose to train with the RoBERTa encoder, classifier and QQP dataset from original MIS study.

We tried various experiment configurations, ranging from the use of new datasets to alterations in architecture and training methods. We will describe all the modifications presented: 

\setlist{nolistsep}
\begin{enumerate}
    \item \textbf{MIS}: Vanilla MIS from HuggingFace Hub without any fine-tuning.
    \item \textbf{MIS trained with LoRA}: Add LoRA adapters instead of partially unfreezing layers.
    \item \textbf{MIS with Vectara}: Replace the original RoBERTa encoder with Vectara's model.
    \item \textbf{MIS with one encoder}: Change MIS two-folded architecture with a single one.
    \item \textbf{MIS trained on the PAWS}: Add 108,463 human-labeled paraphrase adversaries from PAWS.
    \item \textbf{MIS trained on our synthetic data}: Add our synthetic data obtained previously.
\end{enumerate}

\renewcommand{\thefootnote}{\fnsymbol{footnote}}
\begin{table*}[h]
\small
\begin{center}
\begin{tabular}{@{}lllll@{}}
\toprule
\multicolumn{1}{c}{\multirow{2}{*}{\textbf{Method}}} & \multicolumn{2}{c}{\textbf{val}}   & \multicolumn{2}{c}{\textbf{test}}  \\ \cmidrule(l){2-5} 
\multicolumn{1}{c}{}                                 & \textbf{agnostic} & \textbf{aware} & \textbf{agnostic} & \textbf{aware} \\ \midrule
ahoblitz$^{\ast}$                      & -                 & -              & \textbf{0.85}              & \textbf{0.81}           \\
zackchen$^{\ast}$                                             & -                 & -              & 0.84              & \textbf{0.81}           \\
liuwei$^{\ast}$                                               & -                 & -              & 0.83              & 0.80           \\ \midrule
Voting                                               & 0.85              & 0.82           & \underline{0.82}              & 0.78           \\
Normalized averaging                                 & 0.81              & 0.81           & 0.81              & 0.79           \\
MIS + PAWS                                           & 0.82              & 0.82           & 0.81              & 0.78           \\
SFT E5 Mistral                 & 0.83              & 0.77           & 0.80              & 0.77           \\ \midrule
MIS                                                  & 0.81              & 0.78           & 0.80              & 0.77           \\
E5 Mistral                               & 0.81              & 0.80           & 0.76                & 0.78              \\
Vectara                                              & 0.76              & 0.76           & 0.75                 & 0.77              \\
TrueTeacher                          & 0.79              & 0.79           & 0.76                 & \underline{0.80}              \\
GPT-4                                                & -                 & 0.74           & -                 & -              \\ \midrule
SimCSE                           & 0.80          &  0.80 &  0.76             &  0.76         \\
BLEURT                           & 0.77          &  0.77          &  0.74             &  0.74         \\
LaBSE                            & 0.72          &  0.75          &  0.69             &  0.73         \\
METEOR                           & 0.68          &  0.71          &  0.67             &  0.69         \\
chrF                             & 0.63          &  0.72          &  0.65             &  0.67         \\
BLEU                             & 0.67          &  0.70          &  0.64             &  0.65         \\ \midrule
Official baseline                                             & -                 & -              & 0.70              & 0.74           \\ \bottomrule
\end{tabular}
\caption{Performance of described approaches. Accuracy is observed as evaluation score.\hspace{\textwidth}$^{\ast}$Top approaches from the official rankings.}\label{tab:total-method}
\end{center}
\end{table*}

\subsection{Content Preservation Measures}

We conducted a separate analysis on several NLP techniques as examined in the original MIS study. This exploration aimed to assess their suitability for the task of hallucination detection, considering the inherent connection between style transformation, paraphrase generation, and hallucination detection. A well-executed paraphrase should retain the essence of the original text without introducing extraneous elements, which is particularly crucial given that one of the competition's subtasks involved paraphrasing. Specifically, our investigation involved LaBSE, SimCSE, and the metrics for evaluating content preservation described in Section \ref{sec:related-work}.

\subsection{Ensembling}\label{sec:ens}
To enhance the performance of different pre-trained models, we combined them into an ensemble. The final decision on the presence of hallucinations is based on the predictions of multiple independent models.

The predictions of separate models were normalized so that the decision boundary was the same for all models. Thus, differences in the scale of the threshold value did not introduce bias into the final decision.

We have chosen the best set of models for the ensemble from the possible options: E5-Mistral, fine-tuned E5-Mistral, Vectara, TrueTeacher, \textit{all-mpnet-base-v2}\footnote{\url{https://huggingface.co/sentence-transformers/all-mpnet-base-v2}} and also Mutual Implication Score. We calculated cosine between the encoded representations of the model's hypothesis and the target sentence. To obtain a prediction, this score was compared with a descision boundary. For each model we select the optimal classification threshold on validation subset for each track and task. For Vectara we used a threshold of $0.5$. 

We employed different strategies on aggregating individual hallucination scores: Normalized averaging and Voting.

\subsubsection{Normalized averaging}
The predictions of separate models were normalized so that the decision boundary was the same for all models. Thus, differences in the scale of the threshold value did not introduce bias into the final decision.

Individual model scores are normalized as follows: \[\hat{p} = \begin{cases} kp + b, & p \ge \text{thr} \\ \frac{p}{2\text{thr}},&p < \text{thr}  \end{cases}\]
where $k = \frac{1}{2 (1 - \text{thr})},\;b = 1 - k$ and $\text{thr}$ is the optimal decision boundary on validation. 

This transformation allows to keep the score within $[0, 1]$, at the same time, the decision boundary for all models becomes $0.5$.

\subsubsection{Voting}
Another strategy is to aggregate the binary predictions of the models in an ensemble. The presence of hallucinations was determined by voting models, depending on the number of votes in favor. At the verification stage, we determine the minimum number of model votes required to acknowledge the pair of sentences, model hypothesis and ground truth, as a paraphrase, for example, at least one, two or three models voted in favor. That is, we predicted a hallucination if an insufficient number of models compared to the optimal validation threshold classified the sample as a paraphrase.

\begin{table*}[]
\resizebox{\textwidth}{!}{
\begin{tabular}{@{}llllll@{}}
\toprule
\multirow{2}{*}{\textbf{Method}}      & \multirow{2}{*}{\textbf{Models}}                        & \multicolumn{2}{c}{\textbf{val}}   & \multicolumn{2}{c}{\textbf{test}}  \\ \cmidrule(l){3-6} 
                                      &                                                         & \textbf{agnostic} & \textbf{aware} & \textbf{agnostic} & \textbf{aware} \\ \midrule
\multirow{2}{*}{Voting}               & MIS + E5-Mistral + SFT E5-Mistral + all-mpnet + Vectara & \textbf{0.85}     & \textbf{0.82}  & \textbf{0.82}     & 0.78           \\ \cmidrule(l){2-6} 
                                      & MIS + E5-Mistral + SFT E5-Mistral + all-mpnet           & \textbf{0.85}     & 0.80           & \textbf{0.82}     & 0.77           \\ \midrule
\multirow{2}{*}{Normalized averaging} & MIS + E5-Mistral + SFT E5-Mistral                       & \textbf{0.85}     & 0.79           & 0.81              & 0.78           \\ \cmidrule(l){2-6} 
                                      & MIS + all-mpnet + Vectara + TrueTeacher                 & 0.81              & 0.81           & 0.81              & \textbf{0.79}  \\ \bottomrule
\end{tabular}
}
\caption{Ensembling results. Accuracy is observed as evaluation score.}
\label{tab:ens}
\end{table*}

\section{Results}
The comparative analysis of the performance across all baselines, our proposed methods, and the leading approaches derived from the official rankings is collated in Table \ref{tab:total-method}.

\subsection{Ensembling}
According to the results, the Voting approach we developed surpasses all baselines as well as other methods we devised. Nevertheless, the performance narrowly trails the foremost methods from the model-agnostic track in the official rankings by a minimal margin of 0.01. In regards to the application of Ensembling methods, a detailed evaluation delineating the constituent models employed is documented in Table \ref{tab:ens}. It was discerned that the incorporation of our SFT E5-Mistral model enhances overall performance metrics.

\subsection{MIS}
Succeeding in performance ranking is the MIS model, refined through training on the PAWS dataset. As previously elucidated, an assortment of configurations was examined, the details of which are exhaustively represented in Table \ref{tab:mis-ablation}. It is observed that the original MIS model's performance was not substantially uplifted; modifications yielded no marked increment in accuracy. Nonetheless, it is notable that the integration of the PAWS dataset into the training process marginally amplified accuracy for both tracks.  Simultaneously, a minor enhancement on the aware track was observed upon the deployment of the Vectara encoder in place of the RoBERTa model.

\begin{table}
\resizebox{\columnwidth}{!}{
\begin{tabular}{@{}lllll@{}}
\toprule
\multirow{2}{*}{\textbf{Method}} & \multicolumn{2}{c}{\textbf{val}}   & \multicolumn{2}{c}{\textbf{test}}  \\ \cmidrule(l){2-5} 
                                 & \textbf{agnostic} & \textbf{aware} & \textbf{agnostic} & \textbf{aware} \\ \midrule
MIS (original)                   & 0.80              & 0.78           & 0.77              & \textbf{0.80}           \\ \midrule
+ LoRA                           & 0.79              & 0.79           & 0.78              & \textbf{0.80}           \\
+ Vectara                        & 0.79              & 0.81           & \textbf{0.81}              & 0.77           \\
+ Single fold                    & 0.78              & 0.77           & 0.75              & 0.78           \\
+ PAWS                           & \textbf{0.82}     & \textbf{0.82}  & \textbf{0.81}     & 0.78           \\
+ Synthetic data                 & 0.79              & 0.77           & 0.77              & 0.74           \\ \bottomrule
\end{tabular}
}
\caption{MIS ablation study results. Accuracy is observed as evaluation score.}
\label{tab:mis-ablation}
\end{table}

\subsection{SFT E5-Mistral}
The next approach by performance is our SFT E5-Mistral. The accuracy for different configurations in our synthetic data experiments can be found in Table \ref{tab:e5-synt-data-ablation}. The combination of PG and DM synthetic data achieves the best results. Unexpectedly, the use of synthetic data from GPT-4 does not yield as good outcomes. This suggests that GPT-4's synthetic data may contain some inherent biases. 

We carried out a detailed evaluation of a particular subset and identified probable causes for bias:
\begin{itemize}
    \item For texts generated without hallucinations, they tend to be overly formal and intricate.
    \item In cases with hallucinations, numerous instances are exceedingly convoluted, sometimes to the extent that the sentences convey the opposite meaning. Our investigation revealed that such hallucinations might not be readily detectable.
\end{itemize}
It is also clear that relying solely on DM synthetic data does not sufficiently address other tasks. By contrast, a model checkpoint trained with PG synthetic data shows promising performance. Just like the MIS approach, it appears that having PG data is sufficient to address hallucinations in other tasks, provided that the target is accessible.

\begin{table}
\begin{center}
\small
\begin{tabular}{@{}llll@{}}
\toprule
\textbf{Source}        & \textbf{Subset} & \textbf{agnostic}  & \textbf{aware} \\ \midrule
GPT                    & PG              & 0.76          & 0.72           \\ \midrule
\multirow{7}{*}{LLaMA} & PG              & 0.81          & 0.75           \\
                       & DM              & 0.63          & 0.51           \\
                       & MT              & 0.79          & 0.71           \\
                       & PG + DM         & \textbf{0.83} & \textbf{0.77}  \\
                       & PG + MT         & 0.81          & 0.76           \\
                       & MT + DM         & 0.75          & 0.71           \\
                       & All             & 0.77          & 0.71           \\ \midrule
GPT + LLaMA            & All             & 0.77          & 0.73           \\ \bottomrule
\end{tabular}
\end{center}
\caption{Synthetic data ablation study on E5-Mistral. Accuracy is observed as evaluation score.}
\label{tab:e5-synt-data-ablation}
\end{table}

\subsection{Black-box baselines}
All our advanced methods outperform black-box baselines on model-agnostic track. Even though, we observe that the E5-Mistral and MIS methods sets a solid baseline on model-agnostic track, maintaining a high level of performance even without any fine-tuning. Considering model-aware track, all baseline models except of GPT-4 show similar performance. The GPT-4 model  does not do as well as the others in terms of the average score with our specific prompts. Finally, there is the official baseline that our approaches outperform.

\subsection{Content Preservation Measures}

Across preservation measures, SimCSE demonstrates the most notable results. In the model-agnostic track, it performs at the same level as more sophisticated approaches such as TrueTeacher, Vectara, or E5 Mistral, without any fine-tuning. However, other preservation measures do not perform as well. Most of them, with the exception of BLEURT, perform even worse than the official baseline in the model-agnostic track.

\section{Conclusion}
We conducted a comparative analysis involving six baseline models (MIS, E5-Mistral, Vectara, TrueTeacher, GPT-4, and the official baseline from the participant kit) alongside four sophisticated approaches (Voting and Normalized Averaging in Ensembling, as well as the refined MIS and SFT E5-Mistral). Of all methods evaluated, Ensembling demonstrated the highest performance. Nonetheless, the refined MIS and the SFT E5-Mistral exhibited only a minor shortfall in performance when compared to these leading methodologies.

Indeed, there appear to be several avenues for enhancing our synthetic data to potentially exceed the performance of other methods:

\begin{itemize}
    \item Instead of training separate adapters for each task, centralized training with one adapter across multiple tasks could enrich the learning context and expand the size of the training dataset.
    \item Exploring a range of other models, such as Mistral-7b \cite{jiang2023mistral}, Mixtral-8x7b\footnote{\url{https://huggingface.co/mistralai/Mixtral-8x7B-v0.1}}, or LLaMA models of various larger sizes (LLaMA-13b, LLaMA-30b), could identify more efficient architectures or models that are better suited to handle the synthetic data effectively.
    \item For improving the quality of GPT-generated synthetic data, incorporating a more extensive range of examples within few-shot prompts and providing detailed explanations for the \textit{correct} samples could help in mitigating bias and increasing the fidelity of the generated data.
\end{itemize}

The potential use of our adapters to generate both positive and negative samples aimed at a specific target is indeed promising. By assembling datasets that offer these contrasting examples, we could refine the training process through contrastive fine-tuning. Such a method is hypothesized to yield superior performance by facilitating the model's ability to discern and learn from the nuanced differences between correct and incorrect instances.

\bibliography{custom}

\onecolumn

\appendix


\section{GPT-4 prompt for PG task evaluation}
\label{sec:appendix-gpt4-prompt-annot}
\begin{figure*}[ht]
\begin{small}
\begin{spverbatim}
Read the source sentence and the paraphrased hypothesis and answer whether there are any hallucinations or related observable overgeneration errors for the paraphrasing task. 
Before answering, think step by step and write why you chose the answer you did. 
Answer the last string with 'The hypothesis is correct' if there are no hallucinations or misgenerations. Otherwise, answer with 'The hypothesis is false'.

Example 1:
Source sentence: "The European Parliament does not approve the budget."
Paraphrased hypothesis: "The budget cannot be adopted against the will of the European Parliament."
The hypothesis is false

Example 2:
Source sentence: "Everyone is capable of enjoying a good education in a society."
Paraphrased hypothesis: "We must create a society where everyone is able to enjoy a good education."
The hypothesis is correct
\end{spverbatim}
\end{small}
\caption{Prompt for GPT-4 evaluation on PG task.}
\end{figure*}

\newpage

\section{GPT-4 prompt for synthetic paraphrased data generation with hallucinations}
\label{sec:appendix-gpt-synthetic-data-prompt}

\begin{figure*}[ht]
\begin{small}
\begin{spverbatim}
Your aim is to produce an incorrectly paraphrased sentence that contains a hallucination for the given source sentence. Hallucinations in a paraphrase can add new information that wasn't present in the source sentence, or exclude some important information, or reverse the meaning of the source sentence. Remember that reversing source sentence has the lowest level of priority, so use it only if there is no other way to make a hallucination. Usually it's much better to misrepresent some information, add new or exclude something important. If there is some quantitative information in the source, feel free to change them slightly. Complete the task using the examples below. The examples also show the correct paraphrase for the source sentences. Note that there are no hallucinations in the correct paraphrase, whereas your aim is to corrupt the source and produce a false paraphrase. 

Examples:
Source: "I have a permit."
The correct paraphrase: "Uh, I’m validated."
The incorrect paraphrase: "I have a permit to carry it."
Explanation: The incorrect paraphrase adds information that is not present in the source sentence ("to carry it")

Source: "Easy, easy."
The correct paraphrase: "Watch it now."
The incorrect paraphrase: "The process is easy."
Explanation: The incorrect paraphrase introduces additional information ("The process is") 

Source: "A five, six, seven, eight."
The correct paraphrase: "And 5, 6, 7, 8."
The incorrect paraphrase: "A number between five and eight."
Explanation: While the source sentence is a rhythmic count or sequence of specific numbers, the incorrect paraphrase generalizes it to "a number between five and eight".

Source: "A lot safer that way."
The correct paraphrase: "Because it’s safer."
The incorrect paraphrase: "That is a safer way to travel."
Explanation: The major hallucination lies in the addition of "That is," which wasn't present in the original source sentence. This introduces a new element and changes the focus from the general concept of safety to a specific way of travel

Source: "You’re a scam artist."
The correct paraphrase: "You are an imposter."
The incorrect paraphrase: "You’re not a good scam artist."
Explanation: While the source sentence simply states "You’re a scam artist," the incorrect paraphrase implies a judgment on the person's skill as a scam artist

Don't answer now, read the source and think step by step how to make a false paraphrase for the source sentence. Before answering, provide several examples with explanations and choose the best one. Answer starting with 'The incorrect paraphrase: 
\end{spverbatim}
\end{small}
\caption{Prompt for PG data with hallucinations generation using GPT-4.}
\end{figure*}

\newpage

\section{Synthetic data examples} \label{sec:appendix-synthetic-data}

\begin{table*}[ht]
\resizebox{\textwidth}{!}{
\begin{tabular}{@{}llll@{}}
\toprule
\textbf{Task}       & \textbf{Target}                      & \textbf{Hypothesis}                  & \textbf{Label}    \\ \midrule
\multirow{7}{*}{MT} & I know you kissed Tom.               & I know you're going to beat someone. & Hallucination     \\
                    & I'll talk to Tom today.              & I'll talk to John tomorrow.          & Hallucination     \\
                    & I should've waited.      & I'll be better tomorrow.            & Hallucination     \\
                    & He was mistaken for his younger brother  & He was taken for his brother.          & Not Hallucination     \\
                    & She passed the examination.          & She successfully completed the certification.          & Not Hallucination     \\
                    & A plane popped up out of nowhere.    & A plane just came out of the blue.   & Not Hallucination \\
                    & No one has seen Tom since.           & No one has seen Tom for a long time. & Not Hallucination \\ \midrule
\multirow{7}{*}{PG} & Wanna talk?                          & Would you want to talk with me?      & Not Hallucination \\
                    & We must not allow this to happen.    & We must prevent this from happening. & Not Hallucination \\
                    & Have you found her?                  & Have you seen her?                   & Hallucination     \\
                    & A word of advice.    & Give me some advice. & Not Hallucination \\
                    & Hold your course.    & You're going the wrong way, man. & Hallucination \\
                    & Can I take a message?    & Can I take a message for you, & Not Hallucination \\
                    & My job?                              & My job is to carry out the trash.    & Hallucination     \\ \midrule
\multirow{7}{*}{DM} & Delicious .                          & (scrambley) A scrambley dish.        & Hallucination     \\
                    & To increase the level or amount of . & To increase in volume.               & Not Hallucination \\
    
                    & Causing the air to be hot .        & Hot. Something that is hot.                    & Not Hallucination \\
                    & (slang, derogatory) schizoid, schizophrenic; crazy        & (transitive) Crazy      & Not Hallucination \\
                    & Covered with petals or petal-like objects.        & planted.                        & Hallucination \\

                    & Alternative form of midstream        & Middle stream                        & Not Hallucination \\
                    & To require                           & take time to finish something.       & Hallucination     \\ \bottomrule
\end{tabular}
}
\caption{Sample of synthetic data generated using LLaMA2-7B}
\end{table*}

\begin{table*}[ht]
\resizebox{\textwidth}{!}{
\begin{tabular}{@{}lll@{}}
\toprule
\textbf{Target}                                                                 & \textbf{Hypothesis}                                                                              & \textbf{Label}    \\ \midrule
That cannot be in our interest!                                                 & It's not beneficial for us!                                                                      & Not hallucination \\
The written language should be made more user-friendly.                         & The spoken language should be made more user-friendly.                                           & Hallucination     \\
I do not think that is quite what the agreement is.                             & I do not think that's the contract we signed.                                                    & Hallucination     \\
The vote will take place tomorrow at 11.30 a.m.                                 & Tomorrow, the voting process is scheduled for 11.30 in the morning.                              & Not hallucination \\
Mrs Green, you have the floor.                                                  & Mrs. Green, you own the flooring.                                                                & Hallucination     \\
I was also in a northern industrial suburb in Milan.                            & I too have been to one of Milan's northern industrial neighborhoods.                             & Not hallucination \\
Mr President, I should like to make a further remark.                           & Mr. President, I would like to add another comment.                                              & Not hallucination \\
Mrs Bonino tells me that no response is necessary.                              & Mrs. Bonino informed me a response isn't required.                                               & Not hallucination \\ \bottomrule
\end{tabular}
}
\caption{Sample of synthetic data generated using GPT-4}
\end{table*}

\end{document}